\documentclass[10pt,journal,compsoc]{IEEEtran}

\newcommand{\ie}{{\em i.e.}}
\newcommand{\eg}{{\em e.g.}}

\newcommand{\vs}{{\em vs}}

\newcommand{\Eq}[1]{Eq. (\ref{#1})}

\newcommand{\Sect}[1]{Section \ref{#1}}

\usepackage{color}
\definecolor{red}{rgb}{1.00,0.20,0.20}
\definecolor{blue}{rgb}{0.20,0.20,1.00}
\definecolor{green}{rgb}{0.00,1.00,0.00}

\usepackage{array}
\newcolumntype{L}[1]{>{\raggedright\let\newline\\\arraybackslash\hspace{0pt}}m{#1}}
\newcolumntype{C}[1]{>{\centering\let\newline\\\arraybackslash\hspace{0pt}}m{#1}}
\newcolumntype{R}[1]{>{\raggedleft\let\newline\\\arraybackslash\hspace{0pt}}m{#1}}

\newcommand{\UseShortAcronyms}{}

\ifdefined\UseShortAcronyms

\else

\fi

\usepackage{cite}
\usepackage{amsmath,amssymb,amsfonts}
\usepackage{algorithmic}
\usepackage{graphicx}
\usepackage{textcomp}

\usepackage{times}
\usepackage{epsfig}
\usepackage{graphicx}
\usepackage{amsmath}
\usepackage{amssymb}

\usepackage{epsfig}
\usepackage{subcaption}
\usepackage{array,multirow}
\usepackage{float}
\usepackage{algorithm2e}

\usepackage{comment}
\usepackage{xcolor}
\usepackage{hyperref}

\usepackage{soul}

\newcommand{\rot}[1]{\parbox[t]{2mm}{\multirow{4}{*}{\rotatebox[origin=c]{90}{#1}}}}

\begin{document}

\title{Self-supervised Domain Adaptation for Computer Vision Tasks}

\author{Jiaolong~Xu,
        Liang~Xiao,
        and~Antonio~M. L\'opez,~\IEEEmembership{Member,~IEEE}
}





\IEEEtitleabstractindextext{%
\begin{abstract}
Recent progress of self-supervised visual representation learning has achieved remarkable success on many challenging computer vision benchmarks. However, whether these techniques can be used for domain adaptation has not been explored. In this work, we propose a generic method for self-supervised domain adaptation, using object recognition and semantic segmentation of urban scenes as use cases. Focusing on simple pretext/auxiliary tasks (e.g. image rotation prediction), we assess different learning strategies to improve domain adaptation effectiveness by self-supervision. Additionally, we propose two complementary strategies to further boost the domain adaptation accuracy on semantic segmentation within our method, consisting of prediction layer alignment and batch normalization calibration. The experimental results show adaptation levels comparable to most studied domain adaptation methods, thus, bringing self-supervision as a new alternative for reaching domain adaptation. The code is available at \href{https://github.com/Jiaolong/self-supervised-da}{this link}\footnote{https://github.com/Jiaolong/self-supervised-da}.
\end{abstract}

}


\maketitle

\IEEEdisplaynontitleabstractindextext

\IEEEpeerreviewmaketitle

\section{Introduction}

Since supervised (deep) machine learning became the key to solve computer vision tasks, the availability of task ground truth ({\ie} supervision information) associated to the raw data ({\ie} images and videos) has been a major practical problem. Training an image or video classifier requires to associate some class or attributes to the whole image/video \cite{Deng:2009, Kuehne:2011, krizhevsky:2012, Soomro:2012}, training an object detector requires manual drawing of object bounding boxes \cite{Everingham:2010, Lin:2014}, training a CNN for semantic segmentation requires the delineation of the borders between the considered classes \cite{Cordts:2016, mapillary:2017}, etc. This kind of ground truth (bounding boxes, class borders) is usually provided by human labeling, which is a costly process prone to errors due to subjectivity and fatigue. Therefore, procedures aiming at reducing human labeling became a research topic in itself too; or alternatively obtaining the most from a fixed budget for new labels. This underlying aim appears under different names depending on the practical situation at hand, {\ie} the learning conditions. Under this umbrella we find concepts such as active learning, self-labeling, transfer learning, domain adaptation, and self-supervision. 

\begin{figure}
\begin{minipage}{\linewidth}
  \centering
  \centerline{\includegraphics[width=\linewidth]{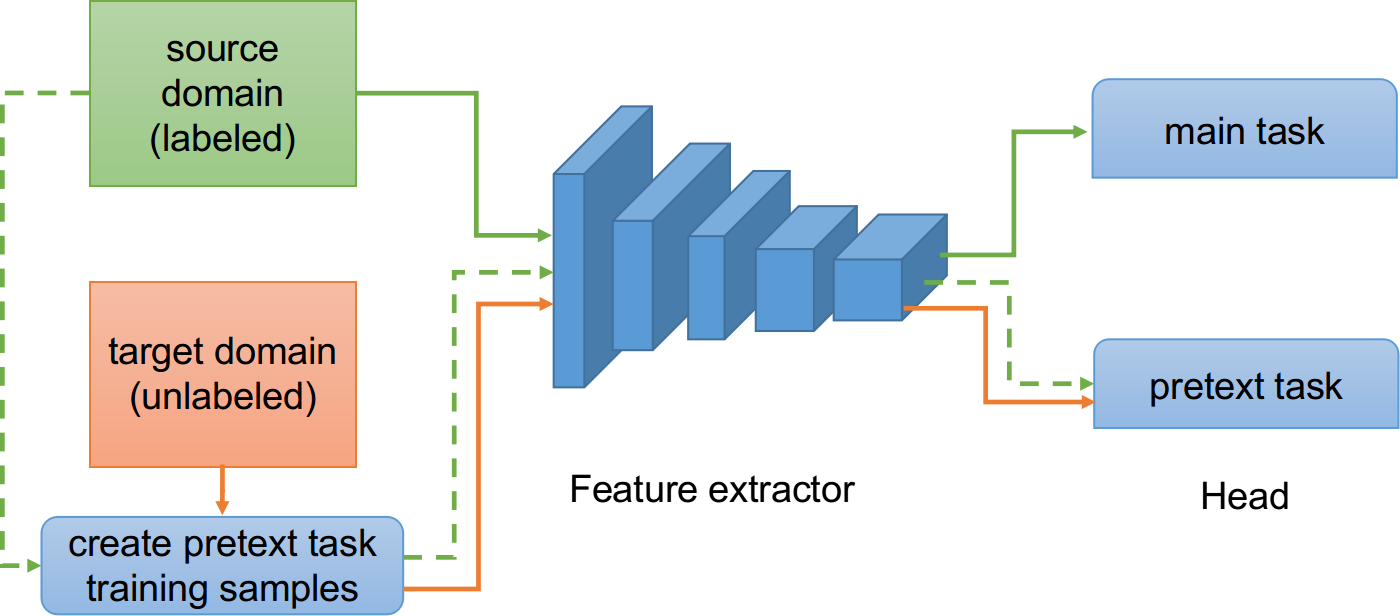}}
\end{minipage}
\caption{Proposed self-supervised domain adaptation framework. We learn a domain invariant feature representation by incorporating a pretext learning task which can automatically create labels from target domain images. The pretext and main task ({\eg} object recognition or semantic segmentation) are learned jointly via multi-task learning. Solid lines indicate the forwarded data flow and the dash lines indicate optional data flow.}
\label{fig:framework}
\end{figure}

In \emph{active learning} \cite{Abramson:2005, settles2012active, Roy2018DeepAL}, the learner receives a set of unlabeled data (videos) for training a visual accurate model, which must be done minimizing the labeling effort by choosing the best training data out of the total amount. This turns out into an iterative process where a human worker labels new automatically selected data in each cycle for model refinement. This contrasts with passive learning, where the training data is selected at random, eventually requiring more labeling budget. 

In \emph{self-labeling} \cite{Xu_PAMI:2014,Xu_ICRA:2016,zou:2018}, an initial visual model is trained on labeled data, after, the model is applied on unlabeled data to self-collect samples which are used then for refining the model by assuming that their label corresponds to the prediction of the model; turning out in an iterative process that must avoid drifting to systematic errors or easy samples. 

In \emph{transfer learning} \cite{Pan:2009,Ross:2014}, a model is trained to perform a visual task ({\eg} image classification) but aiming at reusing it to perform a new task ({\eg} object detection) in a way that we minimize the amount of labeled data required to train for the new task ({\eg} fine-tuning CNNs across tasks is a basic form of transfer learning). 

In \emph{domain adaptation} \cite{fcn_wild:2016, curriculum:2017, chen:2018, da_survey:2018}, a model is trained to perform a visual task in a specific domain ({\eg} semantic segmentation in synthetic images), however, we need to apply it to perform the same task in a correlated, but significantly different,  domain ({\eg} semantic segmentation in real-world images); which is done by reusing the previous knowledge (in the form of model or labeled data) for minimizing the labeling effort in the new domain. 

Finally, \emph{self-supervised learning} \cite{gidaris:2018,jigsaw:2018,revisiting:2019} focuses on learning visual models without manual labeling; more specifically, auxiliary relatively simple tasks, known as \emph{pretext tasks} in this context, are created for training a generic visual model in the form of CNN. The supervision consists in modifying the original visual data ({\eg} a set of images) according to known transforms ({\eg} image rotations \cite{gidaris:2018}), training the pretext CNN to predict such transforms; thus, the transforms are the labels/supervision for the pretext task. This pretext CNN is then concatenated with another task-specific CNN. The former acting as generic feature extractor, and the later leveraging such features to create new ones specific for the \emph{main task} of interest. Sometimes, both CNN blocks are fine-tuned \cite{cluster:2018}, and sometimes the pretext CNN block is frozen and only the task-specific CNN block is fine-tuned \cite{revisiting:2019}. Overall, the idea is that we can have a high number of supervised samples for the pretext task and this should compensate for a lower number of manually labeled samples for the main task.

Active learning can be naturally combined with transfer learning or domain adaptation \cite{Vazquez:2014}. Self-labeling can also be combined with transfer learning or domain adaptation \cite{Xu_PAMI:2014}. Self-supervised learning, as usually performed, can be seen as a type of transfer learning (from the pretext task to the main task). What has not be explored, up to the best of our knowledge, is how self-supervised learning can support domain adaptation. This is the main focus of this paper, {\ie} can we incorporate self-supervision to learn domain invariant feature representation? 
{The goal of this work is not to propose new self-supervised learning methods but investigate how existing self-supervised representation learning methods can be used to address domain adaptation problems.} With this aim, we design a multi-task learning method to jointly train pretext and main tasks (Figure \ref{fig:framework}). The pretext task acts as nexus between source and target domains for learning a domain invariant feature representation for the main task. In this way, we have labels for the main task in source domain, but we do not require labels for such task in the target domain. In other words, via self-supervised learning, we perform unsupervised domain adaptation. 

Accordingly, and using object recognition and semantic segmentation of urban scenes as challenging main-task use cases, the main contributions of this work are three-fold:
\begin{itemize}
\item We proposed a generic method for domain adaptation with self-supervised visual representation learning.
\item Focusing on the image rotation prediction pretext learning task, we proposed several variations and studied their domain adaptation performance.
\item We proposed additional strategies to further boost the self-supervised domain adaptation, including prediction layer alignment and batch normalization calibration.
\end{itemize}

This paper is organized as follows. In \Sect{sec:related_work}, we review related self-supervised representation learning and domain adaptation methods. In \Sect{sec:method}, we explain the proposed method. In \Sect{sec:experiment}, we conduct experiments on domain adaptation for object recognition as well as semantic segmentation, via our method. Finally, \Sect{sec:conclusion} summarizes the work and future directions.

\section{Related work}
\label{sec:related_work}

\paragraph{Self-supervised visual representation learning} An extensive review of deep learning-based self-supervised general visual feature learning methods from images or videos is provided in \cite{jing:2019}. The recent work of self-supervised representation learning mainly focus on the design of pretext tasks.  The work of \cite{revisiting:2019} gives a comprehensive study of some state-of-the-art methods. A pretext task of predicting the relative location of image patches was first proposed in \cite{doersch:2015}, where the patch ID is the supervision/label. This initial patch-based method has been followed by several variants \cite{noroozi:2016, jigsaw:2018, mundhenk:2018}. Other works incorporate image colorization \cite{colorful:2016} or image inpainting \cite{pathak:2016} as pretext tasks. Yet other works focus on automatic ways of creating image samples with corresponding labels; for instance, in \cite{cluster:2018} the labels are classes derived from unsupervised image clustering, and in \cite{gidaris:2018} the labels are image rotation angles since from an original image four possible rotations were created. As compared in \cite{revisiting:2019}, the rotation prediction based method \cite{gidaris:2018} has shown promising results for learning high-level image representations. The rotation based method is further improved in \cite{feng:2019} by decoupling  rotation related and unrelated features. Therefore, in this work, we employ this pretext task as well as the location of image patches in line with \cite{doersch:2015}. In \cite{jiang:2018}, relative depth prediction is used as a self-supervised proxy task, which has shown improvements to the downstream tasks, including semantic segmentation and car detection. However, it relies on the video data in order to obtain the relative depth.

\paragraph{Unsupervised Domain adaptation} 
There have been numerous domain adaptation methods proposed for object recognition since \cite{Saenko:2010}. After the pioneer work of \cite{fcn_wild:2016, curriculum:2017}, semantic segmentation has also aroused increasing interests. Among existing domain adaptation methods, some try to align domains at input level, including GAN-based methods \cite{cycada:2018, BDL:2019, DISE:2019} and image stylization ones \cite{stylization:2018, dcan:2018, fcan:2018}. Some focus at feature level adaptation \cite{fcn_wild:2016, adr:2017, dam:2018, MADA:2018, crdoco:2019}, and others on adapting the output space \cite{mcd:2018, manders:2019, adapt_seg_net:2018, Advent:2019}. According to recent surveys \cite{curriculum:2018, da_survey:2018}, most methods are built on the principle of domain adversarial training \cite{Gani:2015}, with differences on how to incorporate it to the training of the segmentation network. Among the adaptation strategies we use as complement to self-supervision, the prediction layer alignment is similar to adversarial training for output space alignment. 

In \cite{zou:2018}, iterative self-labeling and fine-tuning with spatial urban-scene location priors are used to perform the domain adaptation. In \cite{curriculum:2017}, a curriculum learning style is applied, where super-pixels are computed in source and target domains and their distributions must match as auxiliary task during semantic segmentation training. The use of such auxiliary task is similar in spirit to our multi-task learning approach with pretext tasks as nexus between source and target domains. However, neither our auxiliary tasks nor our complementary adaptation strategies are restricted to semantic segmentation, and they are way simpler than computing super-pixels. Comparing to these work, our method is not specifically designed for semantic segmentation but generic for various computer vision tasks.

In \cite{carlucci:2019}, the self-supervised learning method jigsaw puzzle is used for object recognition domain generalization and adaptation.  As we will see in the experimental section, our method outperforms the jigsaw puzzle based method on both object recognition and semantic segmentation tasks. For semantic segmentation, we compare our results to \cite{adr:2017, curriculum:2017, cycada:2018, stylization:2018, dcan:2018, fcan:2018, dam:2018, adapt_seg_net:2018, zou:2018}. The final semantic segmentation accuracy we obtain in target domain is superior to most of these methods, only behind \cite{zou:2018} which is specific for semantic segmentation, and still not being far apart. Moreover, although it is out of the scope of this paper, our method can be complementary to some of the ones aforementioned, such as those based on adapting the input images via GANs.

\section{Method}
\label{sec:method}

In this section, we first introduce our generic framework of self-supervised domain adaptation. Then, we present the considered pretext tasks. Finally, we introduce domain adaptation steps which complement self-supervision.
\subsection{Self-supervised domain adaptation}
\label{sec:self-supervision}
\subsubsection{Overview of the framework}
\label{sec:overview}
Taking semantic segmentation as an example of main task, but without lose of generality, our method is shown in \figurename~\ref{fig:framework}; where $E$ denotes an encoder network (feature extractor) and $S$ a decoder network (specific of the main task), so that $E+S$ is a CNN for semantic segmentation. This CNN is trained end-to-end with source domain labeled samples, \{$\mbox{X}_{s}$, $\mbox{Y}_{s}$\}. We denote by $P$ the network added to support the creation of a model for solving the pretext task. This model consists in the CNN $E+P$, where $E$ is shared with the CNN of the main task. The pretext task training samples, \{$\mbox{X}_{t}$, $\mbox{Y}_{t}$\}, are automatically created from the target domain images so that the training of $E+P$ is also supervised. 

The complete domain adaptation method is drawn in Algorithm ~\ref{alg:alg1}, where we can see how the self-supervised domain adaptation is a joint training of models to perform the pretext and main tasks. During the forward propagation, both source and target domain samples pass through the shared encoder. After, the losses of the main task $\mathcal{L}_{seg}$ and pretext task $\mathcal{L}_{p}$ are computed, they are back-propagated and accumulated at the encoder. Because the encoder is trained with both source and target domain samples, it learns domain invariant feature representations. In the testing phase, we feed the target domain images to the encoder and pass the features to the decoder of the main task to obtain the predictions. 

\begin{algorithm}
\caption{Self-supervised domain adaptation}
\SetAlgoLined
\KwData {Labeled source domain images: \{$\mbox{X}_{s}$, $\mbox{Y}_{s}$\}, and unlabeled target domain images: \{$\mbox{X}_{t}$\}}

\KwResult{Model trained for main task in target domain}

Create samples for pretext task: \{$\mbox{X}_{t}$, $\mbox{Y}_{t}$\}\;
 $i = 0$\;
 \While{$i < max\_iters$}{
  Load target mini-batch \{$\textbf{x}^t_i$, $\textbf{y}^t_i$\}\;
  Forward pass and compute $\mathcal{L}_{p}$\;
  Back-propagate $\mathcal{L}_{p}$ gradients by $P$ and $E$\;
  Update weights of $P$;
  
  Load source mini-batch \{$\textbf{x}^s_i$, $\textbf{y}^s_i$\}\;
  Forward pass and compute $\mathcal{L}_{seg}$\;
  {Back-propagate} $\mathcal{L}_{seg}$ gradients by $S$ and $E$\; {Accumulate gradients from} $\mathcal{L}_{p}$ and $\mathcal{L}_{seg}$ for $E$\;
  Update weights of $E$ and $S$;
 }
 \label{alg:alg1}
\end{algorithm}

It is also possible to create pretext task samples with the source domain data, {\ie}, dash lines in \figurename~\ref{fig:framework}. In this case, the pretext model can be trained with both source and target domain pretext task samples. We investigate this in Section \ref{sec:experiment}.

\subsubsection{Pretext tasks}

In this section, we first introduce the image rotation prediction pretext task. Inspired by the image-patch based methods \cite{doersch:2015,noroozi:2016, jigsaw:2018, mundhenk:2018}, we also take into account the spatial layout of the image and propose a new pretext task.

\paragraph{Image rotation prediction as pretext task.} We select image rotation prediction as pretext task due to its simplicity and superior performance on visual representation learning to other proposals \cite{gidaris:2018}. Given a set of $N_t$ training images from target domain $\mathit{D}_t = \{\textbf{x}^t_i\}_{i=0}^{N_t}$, similar to \cite{gidaris:2018}, we define the set of geometric transformations as 2D image rotations by $0, 90, 180$ and $270$ degrees. We denote the rotation function by $\mathrm{g}(\textbf{x}^t_i,r), r \in [0, 3]$ rotates image $\textbf{x}^t_i$ by $r*90$ degrees. The geometric transformation prediction model $\mathit{P}$ takes feature map from $\mathit{E}$ as input and outputs a probability distribution over all possible geometric transformations. The self-supervised training objective that the geometric transformation model must learn to solve is:
\begin{equation}
\label{eq:obj_rot}
\min_{\boldsymbol{\theta_{e}}, \boldsymbol{\theta_{p}}} \frac{1}{N_t} \sum_{i=1}^{N_t} \mathcal{L}_{p}(\textbf{x}^t_i, \boldsymbol{\theta_{e}},\boldsymbol{\theta_{p}}),
\end{equation}
where $\boldsymbol{\theta_{e}}$ and $\boldsymbol{\theta_{p}}$ are the parameters of the encoder $\mathit{E}$ and pretext network $\mathit{P}$ respectively, $\mathcal{L}_{p}$ is the loss function defined as:
\begin{equation}
\label{eq:loss_rot}
\mathcal{L}_{p} = -\frac{1}{4}\sum_{r=0}^{3}\log(P(\mathit{E}(g(\textbf{x}^t_i,r),\boldsymbol{\theta_{e}}),\boldsymbol{\theta_{p}})).
\end{equation}

By learning to predict the image orientations, the convolutional neural networks also implicitly learn to localize salient objects in the images, recognize their orientations and object types \cite{gidaris:2018}. Such implicitly learned knowledge contains semantic information of the target domain images which is expected to improve the cross-domain feature representation power of the encoder network. In other words, the pretext task with target domain images helps the encoder to learn domain invariant feature representation, thus, helps to achieve domain adaptation.

The work of \cite{gidaris:2018} uses full images from ImageNet \cite{Deng:2009}. However, the images from a specific domain are usually biased to particular structures or patterns, especially at a full image level. If we train a rotation prediction model with full images, the training process could find a trivial solution and, thus, not being able to learn a domain invariant feature representation. To avoid this problem, we first randomly crop an image patch from the full image and then rotate this patch. In this way, we create more difficult and diverse samples for the pretext task.

\begin{figure}
\begin{minipage}{\linewidth}
  \centering
  \centerline{\includegraphics[width=0.8\linewidth]{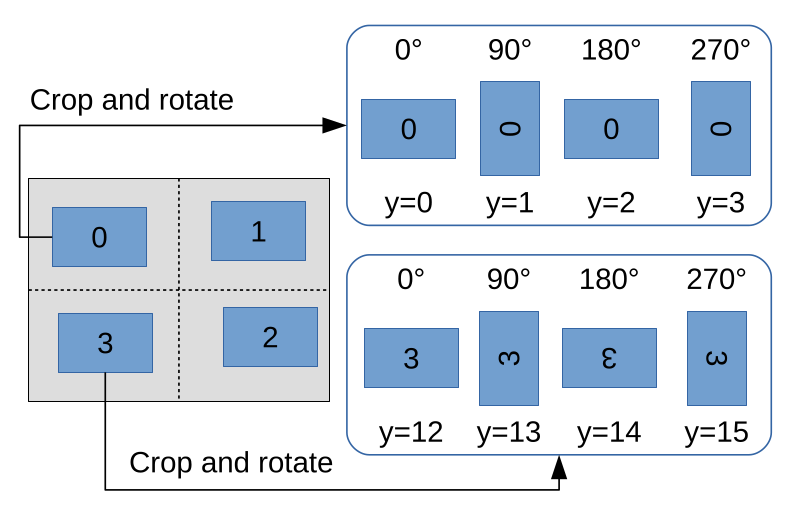}}
\end{minipage}
\caption{Region-based cropping and rotation.}
\label{fig:crop}
\end{figure}
\paragraph{Spatial-aware rotation prediction as pretext task.} Beyond image rotation, we further propose to take into account the image spatial layout to create a more complex pretext task. As depicted in \figurename~\ref{fig:crop}, instead of randomly cropping a patch from the full image, we first split the full image into four regions. From each region, we apply cropping and rotation operations as in the previous pretext task. We call this strategy spatial-aware rotation prediction. The dimension of a label is then extended from $4$ (rotation angles) to $16$ (spatial locations times rotation angles). This scheme encodes the geometry transform as well as spatial layout information, which results in a more complex pretext task.

\subsubsection{Objective function for domain adaptation}
Given a set of $N_s$ labeled training images from the source domain $D_s = \{\textbf{x}^s_i, \textbf{y}^s_i \}_{i=0}^{N_s}$, the segmentation network takes as input the feature maps from $E(\textbf{x}^s_i)$ and outputs the segmentation predictions: $\mathbf{O}^s_i =
S(E(\textbf{x}^s_i,\boldsymbol{\theta_{e}}),\boldsymbol{\theta_{s}}) \in \mathbb{R}^{H \times W \times C}$, where $C$ is the number of semantic categories, $H$ and $W$ are the height and width of the output respectively, and $\boldsymbol{\theta_{e}}$ and $\boldsymbol{\theta_{s}}$ convey the parameters of $E$ and $S$, respectively. The semantic segmentation training objective that we need to solve for $E$ and $S$ is: 
\begin{equation}
\label{eq:obj_seg}
\min_{\boldsymbol{\theta_{e}}, \boldsymbol{\theta_{s}}} \frac{1}{N_s} \sum_{i=1}^{N_s} \mathcal{L}_{seg}(\textbf{x}^s_i, \boldsymbol{\theta_{e}},\boldsymbol{\theta_{s}}), 
\end{equation}
where the segmentation loss is the cross-entropy loss, defined as:
\begin{equation}
\label{eq:loss_seg}
\mathcal{L}_{seg} = -\sum_{h,w}\sum_{c \in C}\textbf{y}^{s}_i(h,w,c)\log(\mathbf{O}^s_i(h,w,c)).
\end{equation}

With \Eq{eq:obj_rot} and \Eq{eq:obj_seg}, the objective function that self-supervised domain adaptation must solve is:
\begin{align}
\label{eq:obj_da}
\min_{\boldsymbol{\theta_{e}}, \boldsymbol{\theta_{p}}, \boldsymbol{\theta_{s}}} &\frac{1}{N_s} \sum_{i=1}^{N_s} \mathcal{L}_{seg}(\textbf{x}^s_i, \boldsymbol{\theta_{e}},\boldsymbol{\theta_{s}}) + \nonumber \\
&\frac{\lambda_p}{N_t} \sum_{j=1}^{N_t} \mathcal{L}_{p}(\textbf{x}^t_j, \boldsymbol{\theta_{e}},\boldsymbol{\theta_{p}}),
\end{align}
where $\lambda_p$ is the weight to balance the two losses. In this work, we simply set $\lambda_p = 1$ for our experiments. The training process follows Algorithm ~\ref{alg:alg1}.
\subsection{Complementary adaptation steps}

In this section, we introduce two different strategies to complement self-supervised domain adaptation, including adversarial training for prediction layer alignment and batch normalization.

\subsubsection{Prediction layer alignment}
\label{sec:adv}
\begin{figure}
\includegraphics[width=\linewidth]{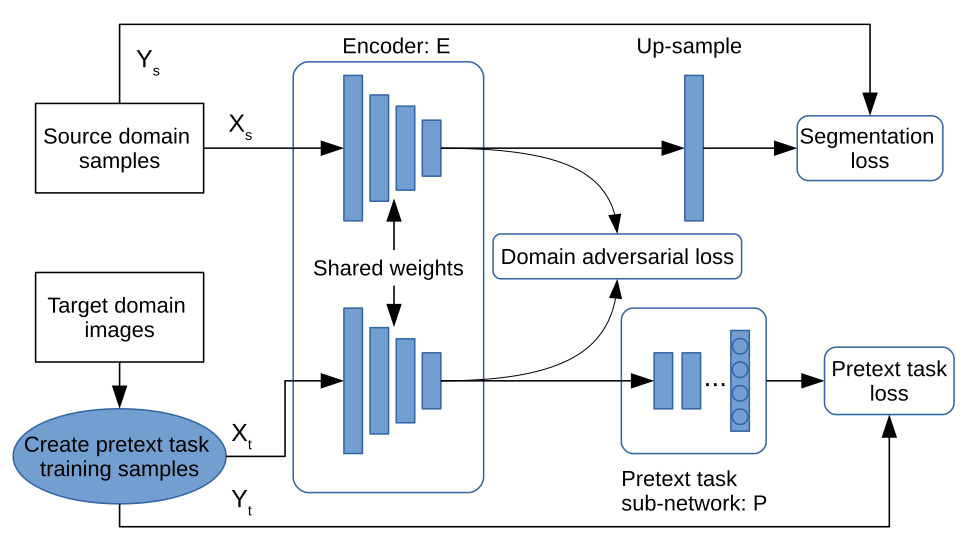}
\caption{Self-supervised domain adaptation with prediction layer alignment.}
\label{fig:rot_adv}
\end{figure}

The proposed pretext task learning is able to perform domain adaptation at feature level, however, the predicted semantic labels may still not be well aligned. There have been some previous work tackling this problem \cite{adapt_seg_net:2018,manders:2019}. In this work, we also consider to align the prediction layer to improve the domain adaptation performance. The main idea is illustrated in \figurename~\ref{fig:rot_adv}. For semantic segmentation, we simplified the decoder by a single up-sampling layer. In this way, the last layer of the encoder is corresponding to the prediction layer. By placing a domain discriminator after the prediction layer, the commonly used domain adversarial training can be employed. We denote by $\mathit{D}$ the discriminator and $\boldsymbol{\theta}_d$ for its parameters. Given an input image $\textbf{x}_i$, the discriminator takes as input the feature maps from the encoder $\mathit{E}(\textbf{x}_i)$ and performs the binary classification to distinguish whether the feature map is from the source image or the target one, $\mathbf{Z}_i = \mathit{D}( \mathit{E}(\textbf{x}_i))$, $\mathbf{Z}_i \in \mathbb{R}^{H \times W \times 2}$. The training of $\mathit{D}$ is a standard supervised training, which minimizes the following 2-D cross-entropy loss:
\begin{align}
\label{eq:l_d}
    \mathcal{L}_{d}(\textbf{x}_i, \boldsymbol{\theta}_d) = - \sum_{h,w} &[(1 - z)\log{\mathbf{Z}_i(h,w,0)} \nonumber \\
    &+ z\log{\mathbf{Z}_i(h,w,1)}],
\end{align}
\noindent where $h,w$ are indexing the output layer, $z = 0$ indicates that the sample is drawn from the target domain, and $z = 1$ if it is drawn from the source domain. 

In order to learn a domain invariant feature representation, we want the encoder to \emph{fool} the domain discriminator $\mathit{D}$, which is equivalent to minimize the following adversarial loss function: 
\begin{align}
\label{eq:l_adv}
    \mathcal{L}_{adv}(\textbf{x}_i, \boldsymbol{\theta}_e) = - \sum_{h,w}&[(1 - z)\log{\mathbf{Z}_i(h,w,1)} \nonumber \\
    &+ z\log{\mathbf{Z}_i(h,w,0)}].
\end{align}
{$\mathcal{L}_{adv}$ encourages to fool $\mathit{D}$ by optimizing $\boldsymbol{\theta}_e$ while $\mathcal{L}_{d}$ encourages to improve the classification accuracy of $\mathit{D}$ by optimizing $\boldsymbol{\theta}_d$}. The optimization of  Eq. (\ref{eq:l_d}) and Eq. (\ref{eq:l_adv}) is essentialy a domain adversarial training. Combining the self-supervised domain adaptation objective function Eq. (\ref{eq:obj_da}), the overall optimization problem that we solve is as following:
\begin{align}
\label{eq:obj_da_adv}
\min_{\boldsymbol{\theta_{e}}, \boldsymbol{\theta_{p}},  \boldsymbol{\theta_{s}}, \boldsymbol{\theta_{d}}} &\frac{1}{N_s} \sum_{i=1}^{N_s} \mathcal{L}_{seg}(\textbf{x}^s_i, \boldsymbol{\theta_{e}},\boldsymbol{\theta_{s}}) \nonumber \\
&+ \frac{\lambda_p}{N_t} \sum_{i=1}^{N_t} \mathcal{L}_{p}(\textbf{x}^t_i, \boldsymbol{\theta_{e}},\boldsymbol{\theta_{p}}) \nonumber \\
&+ \frac{\lambda_{adv}}{N_t + N_s} \sum_{i=1}^{N_t + N_s}\mathcal{L}_{adv}(\textbf{x}_i, \boldsymbol{\theta_{e}}) \nonumber \\
&+ \frac{\lambda_{d}}{N_t + N_s} \sum_{i=1}^{N_t + N_s}\mathcal{L}_{d}(\textbf{x}_i, \boldsymbol{\theta_{d}}),
\end{align}
\noindent where $\lambda_{adv}$ and $\lambda_{d}$ are the weights to balance the corresponding losses. {These hyper-parameters are tuned on the validation set and then fixed for all experiments.} In this work, we set $\lambda_{adv} = 0.01$ and $\lambda_{d} = 1.0$ for our experiments. We show how this prediction layer alignment improves the self-supervised domain adaptation in \Sect{sec:eval_comp}.

\subsubsection{Batch normalization calibration}
\label{sec:bn}
The batch normalization (BN) is originally designed to reduce the internal covariate shift and speedup the training of deep neural networks. Given a mini-batch $\mathcal{B} = \{\textbf{z}_{1\dots m}\}$ as input, BN layer first calculates the mean and variance by $\boldsymbol{\mu}_{\mathcal{B}} = \frac{1}{m}\sum_{i=1}^{m}\textbf{z}_i$, $\boldsymbol{\sigma}^{2} = \frac{1}{m}\sum_{i=1}^{m}(\textbf{z}_i - \boldsymbol{\mu}_{\mathcal{B}})^2$. Each example is then normalized by $\hat{\textbf{z}}_i = \frac{ \textbf{z}_i - \boldsymbol{\mu}_{\mathcal{B}}}{\sqrt{\boldsymbol{\sigma}^{2} + \epsilon}}$, where $\epsilon$ is a constant added to the mini-batch variance for numerical stability. The normalized values are then scaled and shifted by $\lambda \hat{\textbf{z}}_i + \beta$ to produce the output, where $\lambda$ and $\beta$ are learnable parameters.

For a trained source domain model, $\boldsymbol{\mu}_{\mathcal{B}}$ and $\boldsymbol{\sigma}^{2}$ are statistics from source domain images, which may cause domain shift when applied with target domain images. Although during domain adaptation training, both of source and target data are passed through the BN layers, the statistics from both domain can be still ambiguous for the target domain model. What we proposed in this work is to re-calibrate these statistics to reduce the domain shift. Given a pretrained network, we keep all the learnable parameters fixed and feed forward the target domain training images. During this forward propagation, we re-calculate the mean and variation values of each BN layer. 


Our BN calibration is similar to the AdaBN method \cite{adaBN:2018}. However, AdaBN adopts an online algorithm to estimate the mean and variance, while we simply use the common moving average mean and variance available in existing deep learning frameworks. AdaBN is applied at the inference stage, {\ie} to the testing images, while we use BN calibration as a post training process with target domain training images.

\section{Experiments and results}
\label{sec:experiment}
In this section, we conduct experiments to validate the proposed domain adaptation method for both object recognition and semantic segmentation. 

\subsection{Implementation Details}

We implement the proposed method using the PyTorch framework on a single GTX 1080 Ti GPU with $11$ GB memory. For object recognition, we use the code base of JiGen \cite{carlucci:2019} {\footnote{https://github.com/fmcarlucci/JigenDG}. We use the default hyper-parameters and ResNet-18 and ResNet-50 architectures. The deep networks used in our semantic segmentation experiments are ResNet-101 based DeepLab-v2 \cite{deeplabv2:2018} and dilated residual networks (DRN) \cite{drn:2017}. Specifically, we take the commonly used DRN-26 architecture in order to compare to other state-of-the-art methods. Both networks are initialized with ImageNet \cite{Deng:2009} pretrained weights.

\subsection{Domain adaptation for object recognition}

\begin{table*}[]
    \centering
    \begin{tabular}{c|l}
        Methods & Description \\ \hline
        Ours(Jigsaw) &  Self-supervised domain adaptation with pretext task of solving jigsaw puzzle. \cite{carlucci:2019} \\
        Ours(Rot) & Self-supervised domain adaptation with image rotation prediction pretext task. \\
        Ours(MixRot) & Same as Rot but mixing with source domain samples in the pretext task learning. \\
        Ours(SPRot) & Self-supervised domain adaptation with spatial-aware rotation prediction pretext task.\\
        Ours(Adv) & Adversarial domain adaptation with prediction layer alignment.\\
        Ours(Rot+Adv) & Rot with Adv as complementary strategy.\\
        Ours(Rot+Adv+BN) & Rot with Adv and batch normalization calibration as complementary strategies. \\ \hline
    \end{tabular}
    \caption{The abbreviations of the the evaluated methods.}
    \label{tab:abbr}
\end{table*}

We first evaluate the proposed domain adaptation method with state-of-the-art methods on Office \cite{Saenko:2010} dataset. Office is the most widely used dataset for visual domain adaptation, with $4652$ images and $31$ categories collected from three distinct domains: Amazon (\textbf{A}), Webcam (\textbf{W}) and DSLR (\textbf{D}). We evaluate all methods on six domain adaptation tasks, $\mathbf{A} \longrightarrow \mathbf{W}$, $\mathbf{D} \longrightarrow \mathbf{W}$, $\mathbf{A} \longrightarrow \mathbf{D}$, $\mathbf{D} \longrightarrow \mathbf{A}$ and $\mathbf{W} \longrightarrow \mathbf{A}$. We follow the standard protocols for unsupervised domain adaptation in \cite{JAN:2017}, and use all labeled source examples and all unlabeled target examples. We compare the average classification accuracy based on three random experiments. The abbreviations of the evaluated strategies are listed in Table~\ref{tab:abbr}. The results on Office dataset based on ResNet-50 are reported in Table~\ref{tab:office}. \textbf{Rot} achieves the best average accuracy among all evaluated strategies and reaching the accuracies of state-of-the-art methods. \textbf{MixRot} and \textbf{SPRot} obtain similar accuracies which are very close to \textbf{Rot}. \textbf{MixRot} even outperforms \textbf{Rot} on $\mathbf{D} \longrightarrow \mathbf{W}$ task. Because \textbf{Adv} has very low accuracy comparing to \textbf{Rot}, \textbf{Rot+Adv} does not improve \textbf{Rot}. \textbf{Rot+Adv+BN} shows consistent improvements to \textbf{Rot+Adv}, but still can not outperform the simplest method \textbf{Rot}. 

\begin{table*}
\center
\begin{tabular}{cccccccc}
\hline
Method & ${A}\longrightarrow{W}$ & ${D}\longrightarrow{W}$ & ${W}\longrightarrow{D}$ & ${A}\longrightarrow{D}$ &  ${D}\longrightarrow{A}$ & ${W}\longrightarrow{A}$ & Avg. \\ \hline
ResNet-50 \cite{resnet:2016} &68.4$\pm$0.2 &96.7$\pm$0.1 &99.3$\pm$0.1 &68.9$\pm$0.2 &62.5$\pm$0.3 &60.7$\pm$0.3 &76.1 \\
JDDA \cite{JDDA:2019} &82.6$\pm$0.4 &95.2$\pm$0.2 &99.7$\pm$0.0 &79.8$\pm$0.1 &57.4$\pm$0.0 &66.7$\pm$0.2 &80.2 \\  
DAN \cite{DAN:2015} &80.5$\pm$0.4 &97.1$\pm$0.2 &99.6$\pm$0.1 &78.6$\pm$0.2 &63.6$\pm$0.3 &62.8$\pm$0.2 &80.4 \\
RTN \cite{RTN:2016} &84.5$\pm$0.2 &96.8$\pm$0.1 &99.4$\pm$0.1 &77.5$\pm$0.3 &66.2$\pm$0.2 &64.8$\pm$0.3 &81.6 \\
DANN \cite{Gani:2015} &82.0$\pm$0.4 &96.9$\pm$0.2 &99.1$\pm$0.1 &79.7$\pm$0.4 &68.2$\pm$0.4 &67.4$\pm$0.5 &82.2 \\
ADDA \cite{ADDA:2017} &86.2$\pm$0.5 &96.2$\pm$0.3 &98.4$\pm$0.3 &77.8$\pm$0.3 &69.5$\pm$0.4 &68.9$\pm$0.5 &82.9 \\
JAN \cite{JAN:2017} &85.4$\pm$0.3 &97.4$\pm$0.2 &99.8$\pm$0.2 &84.7$\pm$0.3 &68.6$\pm$0.3 &70.0$\pm$0.4 &84.3 \\
MADA \cite{MADA:2018} &90.0$\pm$0.1 &97.4$\pm$0.1 &99.6$\pm$0.1 &87.8$\pm$0.1 &70.3$\pm$0.3 &66.4$\pm$0.3 &85.2 \\
GTA \cite{GTA:2018} &89.5$\pm$0.5 &97.9$\pm$0.3 &99.8$\pm$0.4 &87.7$\pm$0.5 &\textbf{72.8}$\pm$0.3 &\textbf{71.4}$\pm$0.4 &86.5 \\
CDAN \cite{CDAN:2018} &\textbf{94.1}$\pm$0.1 &\textbf{98.6}$\pm$0.1 &\textbf{100.0}$\pm$0.0 &\textbf{92.9}$\pm$0.2 &71.0$\pm$0.3 &69.3$\pm$0.3 &\textbf{87.7} \\ \hline
Ours(Jigsaw) &86.9$\pm$0.8 &\textbf{98.6}$\pm$0.5 &\textbf{100.0}$\pm$0.0 &82.9$\pm$1.0 &62.9$\pm$1.2 &61.2$\pm$0.7 &82.1
\\
Ours(Rot) &90.1$\pm$0.8 &98.1$\pm$0.3 &\textbf{100.0}$\pm$0.0 &88.6$\pm$0.7 &65.1$\pm$0.8 &65.0$\pm$0.6 &84.5
\\
Ours(MixRot) &88.6$\pm$0.7 &98.0$\pm$0.3 &99.9$\pm$0.1 &86.1$\pm$0.3 &65.7$\pm$1.0 &65.0$\pm$0.7 &83.9
\\
Ours(SPRot) &87.3$\pm$0.6 &\textbf{98.5}$\pm$0.7 &\textbf{100.0}$\pm$0.0 &85.1$\pm$0.7 &65.0$\pm$0.7 &62.8$\pm$0.8 &83.1
\\
Ours(Adv) &80.4$\pm$1.1 &98.4$\pm$0.4 &\textbf{100.0}$\pm$0.0 &80.1$\pm$1.9 &65.1$\pm$0.8 &64.9$\pm$0.5 &81.8
\\
Ours(Rot+Adv) &88.4$\pm$0.7 &97.9$\pm$0.3 &\textbf{100.0}$\pm$0.0 &85.5$\pm$0.7 &67.6$\pm$0.6 &65.4$\pm$0.6 &84.1
\\
Ours(Rot+Adv+BN) &88.6$\pm$0.7 &98.0$\pm$0.3 &\textbf{100.0}$\pm$0.0 &85.7$\pm$0.3 &68.0$\pm$0.6 &65.5$\pm$0.3 &84.3 \\
\hline
\end{tabular}
\caption{Accuracy (\%) on Office dataset (ResNet-50).}
\label{tab:office}
\end{table*}

\begin{figure*}
\centering
\begin{minipage}[t]{0.22\textwidth}
\includegraphics[width=\textwidth]{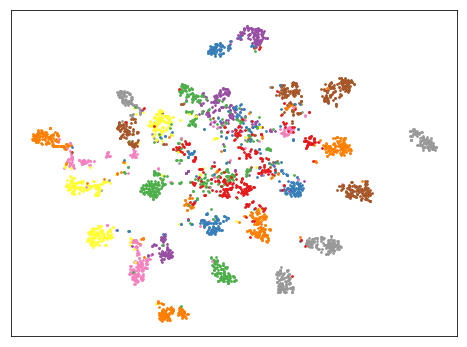}
\subcaption{ResNet-50}
\end{minipage}
\begin{minipage}[t]{0.22\textwidth}
\includegraphics[width=\textwidth]{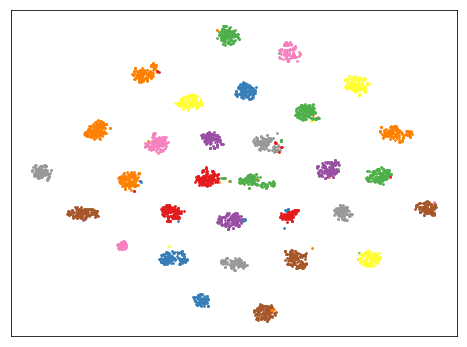}
\subcaption{Rot}
\end{minipage}
\begin{minipage}[t]{0.22\textwidth}
\includegraphics[width=\textwidth]{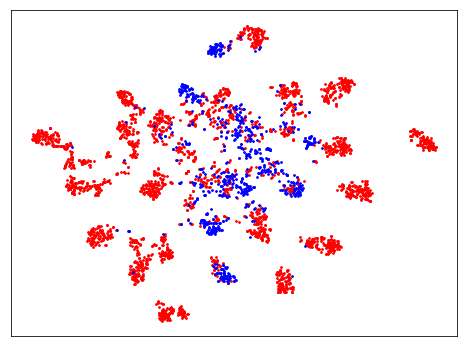}
\subcaption{ResNet-50}
\end{minipage}
\begin{minipage}[t]{0.22\textwidth}
\includegraphics[width=\textwidth]{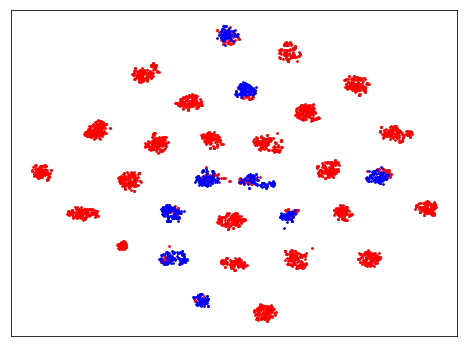}
\subcaption{Rot}
\end{minipage}
\caption{The t-SNE \cite{Maaten:2008} visualization of deep features in $\mathbf{A} \longrightarrow \mathbf{W}$ task. (a)(b) are generated from category information and each color in (a)(b) represents a category. (c)(d) are generated from domain information. Red and blue points represent samples of source and target domains, respectively.}
\label{fig:tsne-office}
\end{figure*}

To better analysis the domain adaptation performance of \textbf{Rot}, we visualize by t-SNE~\cite{Maaten:2008} the learned deep features in \figurename~\ref{fig:tsne-office} on task $\mathbf{A} \longrightarrow \mathbf{W}$. \figurename~\ref{fig:tsne-office} (a) and (b) show that categories are better discriminated by \textbf{Rot} than the non-adapted model ResNet-50. \figurename~\ref{fig:tsne-office} (c) and (d) show that the source and target domains are aligned much better by \textbf{Rot} than ResNet-50. 

For object recognition, we also evaluate on the multiple source domain adaptation dataset PACS \cite{pacs:2017} dataset, which has $7$ object categories and $4$ domains (Photo, Art Paintings, Cartoon and Sketches). \figurename~\ref{fig:pacs} shows sample images from PACS dataset. We follow the same experimental settings as \cite{carlucci:2019} and trained our model considering three domains as source datasets and the remaining one as target. Following \cite{carlucci:2019}, we also compare to the domain discovery method DDiscovery~\cite{mancini:2018} and Dial~\cite{carlucci:2017}. We set three different random seeds and run each experiment three times. The final result is the average over the three repetitions. To make a fair comparison, we run jigsaw puzzle method with the same random seeds and denoted by Ours(jigsaw). The results are shown in Table~\ref{tab:pacs}. We obtained similar conclusions to the Office dataset experiment. Our image rotation based self-supervised domain adaptation \textbf{Rot} outperforms all baselines. \textbf{MixRot} outperforms \textbf{Rot} on adaptation to art painting and photo. \textbf{SPRot} outperforms \textbf{Rot} on adaptation to cartoon and photo. But their overall performance, {\ie}, average accuracies are still lower than \textbf{Rot}, showing that \textbf{Rot} is the most robust method. Again, due to the relatively too low performance of \textbf{Adv}, \textbf{Rot+Adv} can not further improve \textbf{Rot}. As in the previous experiment, \textbf{Rot + Adv + BN} consistantly improves \textbf{Rot + Adv}.

\begin{figure*}
\centering
\includegraphics[width=0.6\linewidth]{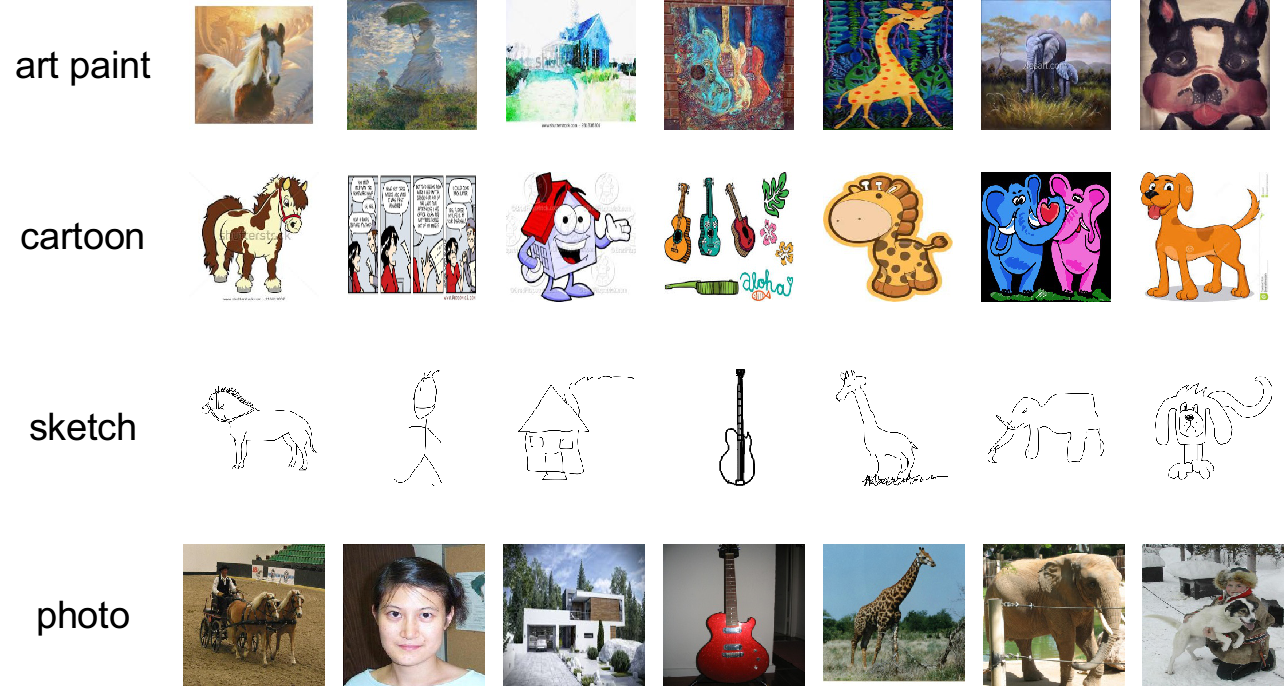}
\caption{Sample images from PACS dataset. Each row represents a domain and each column represents a category.}
\label{fig:pacs}
\end{figure*}

\begin{table}
\center
\begin{tabular}{cccccc}
\hline
Method & art paint. & cartoon & sketches & photo & Avg. \\ \hline
SRC\cite{mancini:2018} & 74.70 & 72.40 & 60.10 & 92.90 & 75.03 \\
Dial\cite{carlucci:2017} & 87.30 & 85.50 & 66.80 & 97.00 & 84.15 \\
DDiscovery\cite{mancini:2018} & 87.70 & 86.90 & 69.60 & 97.00 & 85.30 \\ \hline
SRC\cite{carlucci:2019} & 77.85 & 74.86 & 67.74 & 95.73 & 79.05 \\
JiGen\cite{carlucci:2019} & 84.88 & 81.07 & 79.05 & 97.96 & 85.74 \\ \hline
Ours(SRC) & 79.33 & 76.75 & 64.40 & 96.39 & 79.22 \\
Ours(Jigsaw) & 84.93 & 83.85 & 69.04 & 93.92 & 82.94 \\
Ours(Rot) & 88.67 & 86.39 & \textbf{74.93} & 98.02 & \textbf{87.00} \\
Ours(MixRot) & \textbf{89.35} & 84.14 & 74.49 & \textbf{98.24} & 86.56 \\
Ours(SPRot) & 86.57 & \textbf{87.95} & 67.06 & \textbf{98.26} & 84.96 \\
Ours(Adv) & 80.68 & 77.45 & 67.80 & 96.81 &  80.69 \\
Ours(Rot+Adv) & 86.47 & 86.18 & 70.53 & 97.90 &  85.27 \\
Ours(Rot+Adv+BN) & 86.87 & 86.52 & 71.59 & 98.01 &  85.75 \\
\hline
\end{tabular}
\caption{Multi-source Domain Adaptation results on PACS (ResNet-18).  Three domains are used as source datasets and the remaining one as target.}
\label{tab:pacs}
\end{table}

\begin{figure*}
\centering
\begin{minipage}[t]{0.22\textwidth}
\includegraphics[width=\textwidth]{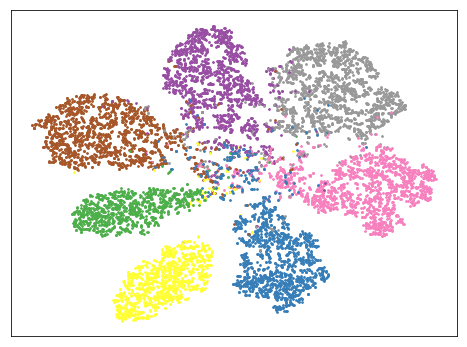}
\subcaption{SRC}
\end{minipage}
\begin{minipage}[t]{0.22\textwidth}
\includegraphics[width=\textwidth]{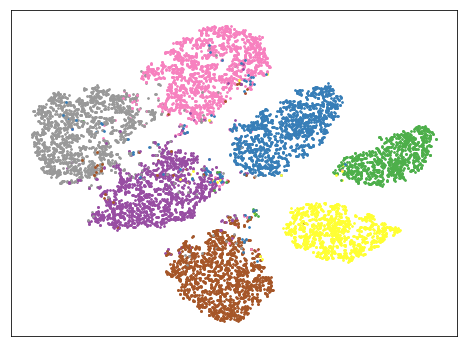}
\subcaption{Rot}
\end{minipage}
\begin{minipage}[t]{0.22\textwidth}
\includegraphics[width=\textwidth]{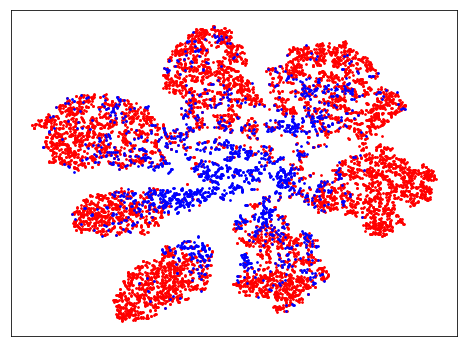}
\subcaption{SRC}
\end{minipage}
\begin{minipage}[t]{0.22\textwidth}
\includegraphics[width=\textwidth]{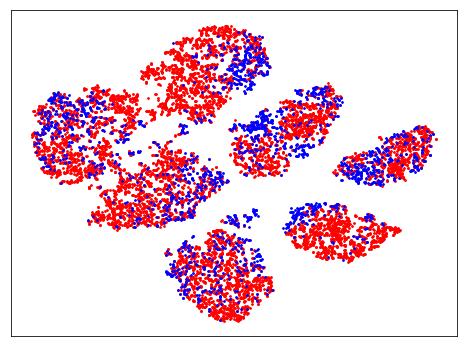}
\subcaption{Rot}
\end{minipage}
\caption{The t-SNE \cite{Maaten:2008} visualization of deep features in multi-source DA task (art painting is used as target domain). (a)(b) are generated from category information and each color in (a)(b) represents a category. (c)(d) are generated from domain information. Red and blue points represent samples of source and target domains, respectively.}
\label{fig:tsne-pacs}
\end{figure*}

For domain adaptation on PACS, we also show t-SNE \cite{Maaten:2008} visualization of deep features in \figurename~\ref{fig:tsne-pacs}. From (a) and (b), we can see that \textbf{Rot} has much better discriminativity on categories than non-adapted method \textbf{SRC}. From (c) and (d), \textbf{Rot} shows clearly much better domain alignment than non-adapted method \textbf{SRC}. The t-SNE visualization reveals the effectiveness of \textbf{Rot} on domain adaptation.

\subsection{Domain adaptation for semantic segmentation}

For semantic segmentation, we adapt semantic segmentation models from the source domain of synthetic images to the target domain of real-world images. For the synthetic datasets, we use SYNTHIA \cite{Ros:CVPR16} and GTA5 \cite{Stephan:ECCV16}, and for the target domain, we use the Cityscapes dataset \cite{Cordts:2016}. The GTA5 \cite{Stephan:ECCV16} dataset is rendered from the Grand Theft Auto V video game. It consists of $24996$ images with resolution of $1914 \times 1052$ and has $19$ classes compatible with Cityscapes dataset. We use the full set of GTA5 as our source domain training set. For SYNTHIA dataset, we use the SYNTHIA-RAND-CITYSCAPES set \cite{Ros:CVPR16} as the source domain training set, which contains $9400$ images. We evaluate with the $16$ common classes for SYNTHIA to Cityscapes domain adaptation. The training set of Cityscapes has $2975$ images which are used as unlabeled target domain training samples. The validation set of Cityscapes has $500$ samples which are used as our testing set.

We conduct ablation studies to understand the impact of each component of our self-supervised domain adaptation. If not otherwise specified, all the experiments in this section use ResNet-101 as backbone network and the domain adaptation is from GTA5 to Cityscapes.

\subsubsection{Pretext task learning strategies}
The first two rows in Table~\ref{tab:ablation} show their domain adaptation results. As can be seen from Table~\ref{tab:ablation}, mixing source domain training data in the pretext learning (\textbf{MixRot}) shows even inferior results, which may because the source samples are dominated in the mixed samples ($24996$ {\vs} $2975$), which makes the model more source domain oriented and reduces domain invariant representation power.

Next, we would like to know whether the proposed spatial-aware rotation prediction pretext task is better than the simple rotation prediction strategy, {\ie}, the \textbf{Rot} method. Table~\ref{tab:ablation} displays the results of the spatial-aware rotation prediction pretext task as \textbf{SPRot}. It turns out that the more difficult pretext task learning leads to worse domain adaptation performance. In our practice, the pretext task learning of \textbf{SPRot} has more difficulties to converge than \textbf{Rot}, and this may result in the failure of learning good feature representations. Therefore, how to design a proper pretext task for domain adaptation still needs more exploration.

\begin{table*}
\center
\scriptsize
\begin{tabular}{m{5mm}m{3mm}m{3mm}m{3mm}m{3mm}m{3mm}m{3mm}m{3mm}m{3mm}m{3mm}m{3mm}m{3mm}m{3mm}m{3mm}m{3mm}m{3mm}m{3mm}m{3mm}m{3mm}m{3mm}m{3mm}m{3mm}}
\hline
\multicolumn{21}{c}{GTA5 $\rightarrow$ Cityscapes}\\ \hline
\multirow{4}{*}{} & \rot{mIoU} & \rot{road}  & \rot{sidewalk} & \rot{building} & \rot{wall} & \rot{fence} & \rot{pole} & \rot{light} & \rot{sign} & \rot{veg} & \rot{terrain} & \rot{sky} & \rot{person} & \rot{rider} & \rot{car} & \rot{truck} & \rot{bus} & \rot{train} & \rot{mbike} & \rot{bike}\\
  \\
  \\
  \\
 \hline
 Rot&\textbf{41.2}&87.6&25.7&77.5&19.8&16.8&29.0&32.1&20.5&79.9&32.9&75.3&58.2&26.0&79.0&23.3&31.6&2.1&26.9&37.7\\
 MixRot&40.3&79.4&13.7&77.6&20.1&19.4&27.9&36.3&30.6&83.3&29.2&74.4&60.2&29.2&64.9&27.8&18.1&0.3&28.4&44.2\\
 SPRot&37.7&80.6&19.2&76.1&17.8&16.0&29.4&32.9&20.2&77.5&19.6&74.2&59.0&28.1&67.8&31.2&12.4&0.4&26.8&25.9\\
    
 \hline
 128x128&38.6&77.8&13.5&78.9&18.6&19.1&25.8&34.3&28.8&77.4&19.2&72.1&60.2&27.3&67.0&31.5&8.3&0.7&32.1&40.7\\
 256x256&\textbf{41.2}& 87.6&25.7&77.5&19.8&16.8&29.0&32.1&20.5&79.9&32.9&75.3&58.2&26.0&79.0&23.3&31.6&2.1&26.9&37.7\\
 400x400&40.3&79.4&13.7&77.6&20.1&19.4&27.9&36.3&30.6&83.3&29.2&74.4&60.2&29.2&64.9&27.8&18.1&0.3&28.4&44.2\\

\hline
Middle&\textbf{41.2}& 87.6&25.7&77.5&19.8&16.8&29.0&32.1&20.5&79.9&32.9&75.3&58.2&26.0&79.0&23.3&31.6&2.1&26.9&37.7\\
Final &40.4&84.1&25.0&79.5&15.5&15.5&29.5&30.5&27.8&82.1&21.7&80.3&54.3&26.0&70.1&29.5&29.2&0.2&26.3&40.9\\

\hline
\end{tabular}
\caption{Domain adaptation performance under different pretext task settings (ResNet-101).}
\label{tab:ablation}
\end{table*}

We also compare our method \textbf{Rot} to the jigsaw puzzle based self-supervision \cite{carlucci:2019}. The results are shown in Table~\ref{tab:jigsaw}, where SYN2CS denotes SYNTHIA to Cityscapes domain adaptation and GTA2CS for GTA5 to Cityscapes. \textbf{Rot} outperforms the jigsaw puzzle for both SYN2CS and GTS2CS. Especially for GTA2CS, jigsaw puzzle has shown very limited gain (1.2 percentage point) while \textbf{Rot} still achieved $6.2$ percentage point.

\begin{table*}
\center
\begin{tabular}{|c||c|c|c|c|c|c|}
\hline
\multirow{2}{*}{Method} & \multicolumn{3}{c|}{SYN2CS} & \multicolumn{3}{c|}{GTA2CS} \\
 & SRC & Adapt & Gain & SRC & Adapt & Gain \\ 

\hline \hline
\multirow{1}{*}{Jigsaw puzzle} & 30.5 & 34.3 & 3.8 & 35.0 & 36.2 & 1.2 \\
\multirow{1}{*}{Ours(Rot)} & 30.5 & 36.1 & 5.6 & 35.0 & 41.2 & 6.2 \\
\hline
\end{tabular}
\caption{Comparison with jigsaw puzzle method (ResNet-101).}
\label{tab:jigsaw}
\end{table*}

\subsubsection{Input image size for pretext task learning}

As the images from Cityscapes dataset have large resolution (\eg, $1024 \times 2048$). We are interested in what cropping size is best for the self-supervised learning. In Table~\ref{tab:ablation}, we compare three different cropping sizes. The smallest cropping size ($128 \times 128)$ shows worst performance due to too small field of view to learn good representations. Comparing the remaining two cropping sizes, we see that the larger one ($400 \times 400$) does not further improve the performance. In fact, when we use the full image as input, the pretext learning easily gets stuck in a trivial solution, {\ie} $100\%$ prediction accuracy. As a result, the final model fails to perform domain adaptation. Thus, we believe that a proper cropping size is important to control the difficulty of learning pretext tasks.

\subsubsection{Feature extraction layer}

By default, the pretext task takes as input the features extracted from the last layer of the encoder. However, whether the last layer is the best for domain adaptation is unclear. In this section, we train self-supervised domain adaption models with different feature extraction layers. We mainly compare the feature extraction from the middle and the end of the encoder. Table~\ref{tab:ablation} shows the corresponding results, where \textbf{Middle} represents the feature extraction from middle layer and \textbf{Final} uses features from the end layer of the encoder. As, in this case, the decoder of the segmentation network is simply an up-sampling layer without any learnable parameter, the \textbf{Final} layer is actually the prediction layer of the segmentation network. As we can see from the results, the model \textbf{Middle} shows slightly better results and we think the pretext task learning is not very sensitive to the choice of feature extraction layers.

\subsubsection{Evaluation of complementary strategies}
\label{sec:eval_comp}
\begin{table*}
\center
\scriptsize
\begin{tabular}{m{7mm}m{3mm}m{3mm}m{3mm}m{3mm}m{3mm}m{3mm}m{3mm}m{3mm}m{3mm}m{3mm}m{3mm}m{3mm}m{3mm}m{3mm}m{3mm}m{3mm}m{3mm}m{3mm}m{3mm}m{3mm}m{3mm}}
\hline
\multicolumn{21}{c}{GTA5 $\rightarrow$ Cityscapes}\\ \hline
\multirow{4}{*}{} & \rot{mIoU} & \rot{road}  & \rot{sidewalk} & \rot{building} & \rot{wall} & \rot{fence} & \rot{pole} & \rot{light} & \rot{sign} & \rot{veg} & \rot{terrain} & \rot{sky} & \rot{person} & \rot{rider} & \rot{car} & \rot{truck} & \rot{bus} & \rot{train} & \rot{mbike} & \rot{bike}\\
  \\
  \\
  \\
 \hline
Rot&41.2&87.6&25.7&77.5&19.8&16.8&29.0&32.1&20.5&79.9&32.9&75.3&58.2&26.0&79.0&23.3&31.6&2.1&26.9&37.7 \\
+Adv&42.3&84.9&31.9&80.4&19.0&21.7&28.2&34.7&27.7&82.8&26.5&72.7&58.2&25.3&82.1&18.7&42.1&1.2&26.0&39.7 \\
+BN&41.0&86.7&32.6&78.7&20.4&20.6&27.0&28.6&15.8&82.4&38.0&74.6&57.6&24.0&80.1&23.6&29.3&0.8&23.2&34.8\\
+Adv+BN&\textbf{43.3}&87.3&35.0&80.0&20.2&21.8&28.7&32.3&25.8&83.3&29.3&73.7&58.7&25.7&83.2&27.5&43.6&1.8&27.5&38.1 \\ \hline
SRC&35.0&77.5&12.3&71.3&8.1&18.8&26.6&32.4&19.6&73.7&11.3&67.9&55.2&24.3&73.3&16.9&9.9&0.9&26.4&39.5 \\
TAR&65.3&96.5&74.3&88.0&48.8&41.2&42.3&47.0&60.7&88.5&52.8&90.5&68.6&48.9&91.1&68.5&69.5&46.3&51.3&65.0 \\
\hline
\end{tabular}
\caption{Evaluation of complementary strategies (ResNet-101).}
\label{tab:complementary}
\end{table*}

\begin{table*}
\center
\begin{tabular}{|c||c|c|c|c|c|c|c|c|}
\hline
\multirow{2}{*}{Method} & \multirow{2}{4em}{Network} & \multicolumn{3}{c|}{SYN2CS} & \multicolumn{3}{c|}{GTA2CS} & \multirow{2}{*}{Mechanism} \\
 & & SRC & Adapt & Gain & SRC & Adapt & Gain & \\ 

\hline \hline





CyCADA\cite{cycada:2018} & DRN-26 & - & - & - & 21.7 & 39.5 & 17.8 & Input adversary \\ 
\hline

Stylization\cite{stylization:2018} & DRN-26 & 22.0 & 35.0 & 13.0 & 22.9 & 38.3 & 15.4 & Input stylization \\ 
\hline

DCAN\cite{dcan:2018} & ResNet-101 & 28.0 & 36.5 & 8.5 & 29.8 & 38.5 & 8.7 & Input stylization \\
\hline

FCAN\cite{fcan:2018} & ResNet-101 & - & - & - & 29.2 & 46.6 & 17.4 & Input stylization + feature adversary\\
\hline

CrDoCo\cite{crdoco:2019} & DRN-26 & 22.9 & 33.4 & 10.5 & 22.9 & 45.1 & 22.2 & Input adversary + feature adversary \\
\hline

DISE\cite{DISE:2019} & ResNet-101 & - & 41.5 & - & - & 45.4 & - & Input adversary + output adversary \\
\hline

BDL\cite{BDL:2019} & ResNet-101 & - & 51.4 & - & 33.6 & 48.5 & 14.9 & Input adversary + self-labelling \\
\hline

ADR~\cite{adr:2017} & ResNet-50 & - & - & - & 25.3 & 33.3 & 8.0 & Feature adversary \\
\hline

GAM\cite{dam:2018} & DRN-26 & - & - & - & - & 40.2 & - & Feature adversary \\ 
\hline


AdaptSegNet\cite{adapt_seg_net:2018} & ResNet-101 & - & - & - & 36.6 & 41.4 & 4.8 & Output adversary\\ 
\hline

ADVENT\cite{Advent:2019} & ResNet-101 & - & 41.2 & - & - & 45.5 & - & Output adversary \\
\hline

CLAN\cite{Luo:2019} & ResNet-101 & 38.6 & 47.8 & 9.2 & 36.6 & 43.2 & 6.6 & Output adversary \\
\hline

CBST\cite{zou:2018} & ResNet-38 & 29.2 & 42.5 & 13.3 & 35.4 & 47.0 & 11.6 & Self-labelling\\
\hline

CURC\cite{curriculum:2017}& DRN-26 & 21.9 & 28.2 & 6.3 & - & - & - & Curriculum\\
\hline

\multirow{2}{*}{Ours(Rot)} & ResNet-101 & 30.5 & 36.1 & 5.6 & 35.0 & 41.2 & 6.2 & \multirow{2}{*}{Self-supervision} \\
& DRN-26 &25.1  &28.9  &3.8  & 29.4 & 34.6 & 5.2 & \\
\hline

\multirow{2}{*}{Ours(Adv)} & ResNet-101 & 30.5 & 35.4 & 4.9 & 35.0 & 40.4 & 5.4 & \multirow{2}{*}{Output adversary} \\
& DRN-26 &25.1 &29.9 & 4.8 & 29.4 & 33.1 & 3.7 &  \\
\hline

\multirow{2}{*}{Ours(Rot+Adv)} & ResNet-101 & 30.5 & 39.3 & 8.8 & 35.0 & 42.3 & 7.3 & Self-supervision \\
& DRN-26 &25.1 &31.7 & 6.6 & 29.4 & 36.1 & 6.7 & + output adversary \\
\hline

\multirow{2}{*}{Ours(Rot+Adv+BN)} & ResNet-101 & 30.5 & 38.8 & 8.3 & 35.0 & 43.3 & 8.3 & Self-supervision \\
& DRN-26 &25.1 & 30.4 & 5.3 & 29.4 & 36.2 & 6.8 & + output adversary + BN\\
\hline

\end{tabular}
\caption{Comparison with the state-of-the-art methods. The result of AdaptSegNet \cite{adapt_seg_net:2018} here is from the single resolution version as our output adversarial method is built on top of this version.}
\label{tab:sota}
\end{table*}

\begin{figure}
\centering
\includegraphics[width=\linewidth]{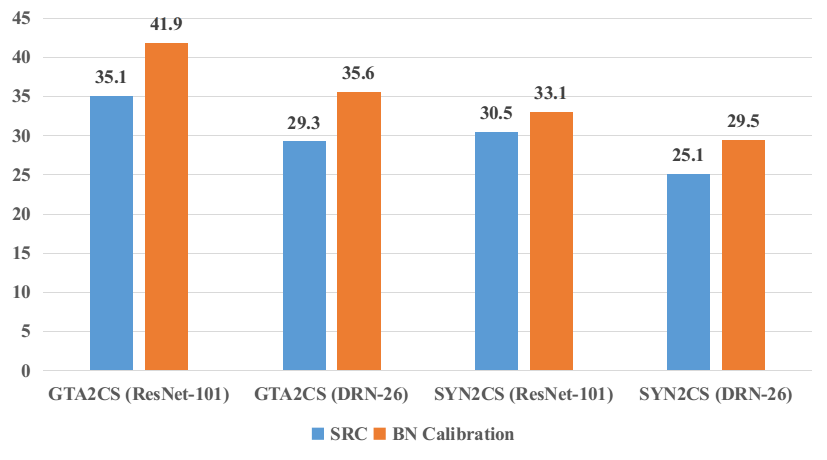}
\caption{Domain adaptation performance of BN calibration. The vertical axis denotes mIoU accuracy.}
\label{fig:bn_cal}
\end{figure}

Table~\ref{tab:complementary} shows the results with different complementary strategies. The source domain model is denoted by \textbf{SRC} and the model trained with target domain samples is denoted by \textbf{TAR}, which represent the lower and upper bound of the accuracy respectively. \textbf{Rot} is our baseline method. \textbf{+Adv} is with prediction layer alignment (\Sect{sec:adv}), which improves \textbf{Rot} by $1.1$ percentage points. \textbf{+BN} is with BN calibration (\Sect{sec:bn}), which does not show improvement over \textbf{Rot}. But when combined \textbf{Adv} and \textbf{BN}, we obtain the best results, improving \textbf{Rot} by $2.1$ percentage points. Tabel~\ref{tab:sota} shows more results with other architectures and datasets, where \textbf{+Adv} has consistent improvements to \textbf{Rot} but \textbf{+Adv+BN} gets saturated for the SYN2CS problem. 

To understand why BN does not have consistent improvements, we further conduct experiments using only BN calibration for domain adaptation. \figurename~\ref{fig:bn_cal} shows the results on multiple datasets using multiple networks. BN calibration alone achieves surprisingly good results, and the best domain adaptation gain even reaches $6.8$ percentage points. However, when combined with \textbf{Rot} or \textbf{Rot+Adv}, it only improves $1$ or $2$ percentage points. This might be because \textbf{Rot} and \textbf{Rot+Adv} have already learned domain invariant representation that effectively reduces the covariate sift and BN calibration could not contribute more to the adapted model. The reason that \textbf{Adv} gives consistent rise to the base method is because \textbf{Adv} further aligns the predicted label distributions which is more complementary adaptation to the \textbf{Rot} than the provided by \textbf{BN}. 

\subsubsection{Qualitative analysis}

\begin{figure*}
\centering
\includegraphics[width=1.0\linewidth]{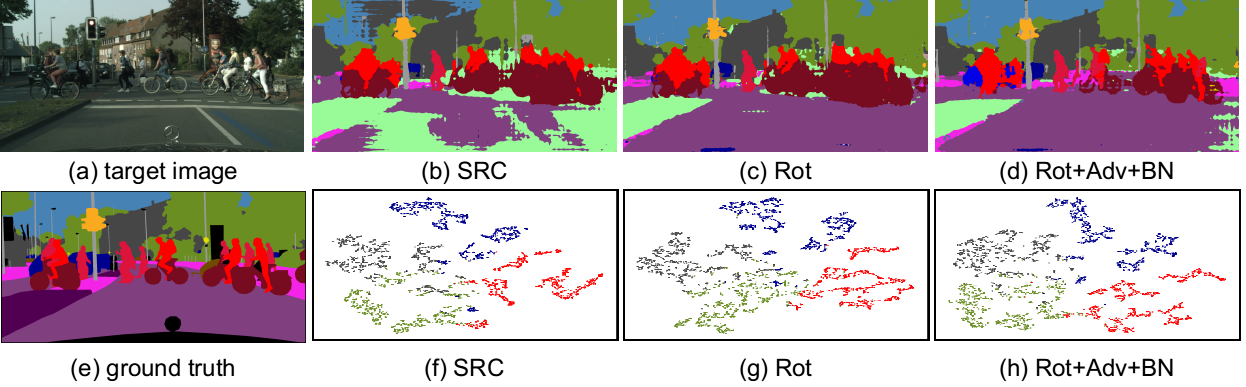}
\caption{The t-SNE \cite{Maaten:2008} visualization of the learned features. (a): A target domain image. (b): A non-adapted results. (c) Adapted result of Rot. (d) Adapted results of Rot+Adv+BN. (e) ground truth segmentation. We map the high-dimensional features of (b), (c) and (d) to a 2-D space with t-SNE \cite{Maaten:2008} shown in (f), (g) and (h). For clear illustration, we visualize features of 4 classes, including building (grey), car (blue), vegetation(green) and rider(red). }
\label{fig:seg-tsne}
\end{figure*}

Following \cite{Luo:2019}, we also visualize the learned feature representations by t-SNE \cite{Maaten:2008} in \figurename~\ref{fig:seg-tsne}. For the non-adapted features, the classes are not discriminated well. They are discriminated better by the \textbf{Rot} adaptation and \textbf{Rot+Adv+BN} discriminates the classes best. In \figurename~\ref{fig:qualitative}, {we illustrate some qualitative results of our models. Without domain adaptation, the source domain model \textbf{SRC} produces noisy segmentation. \textbf{Rot} shows significant improvements over \textbf{SRC} in terms of segmentation quality. The results of \textbf{Rot+Adv+BN} is less noisy and more accurate in details than \textbf{Rot}}. 

\begin{figure*}
\centering
\includegraphics[width=1.0\linewidth]{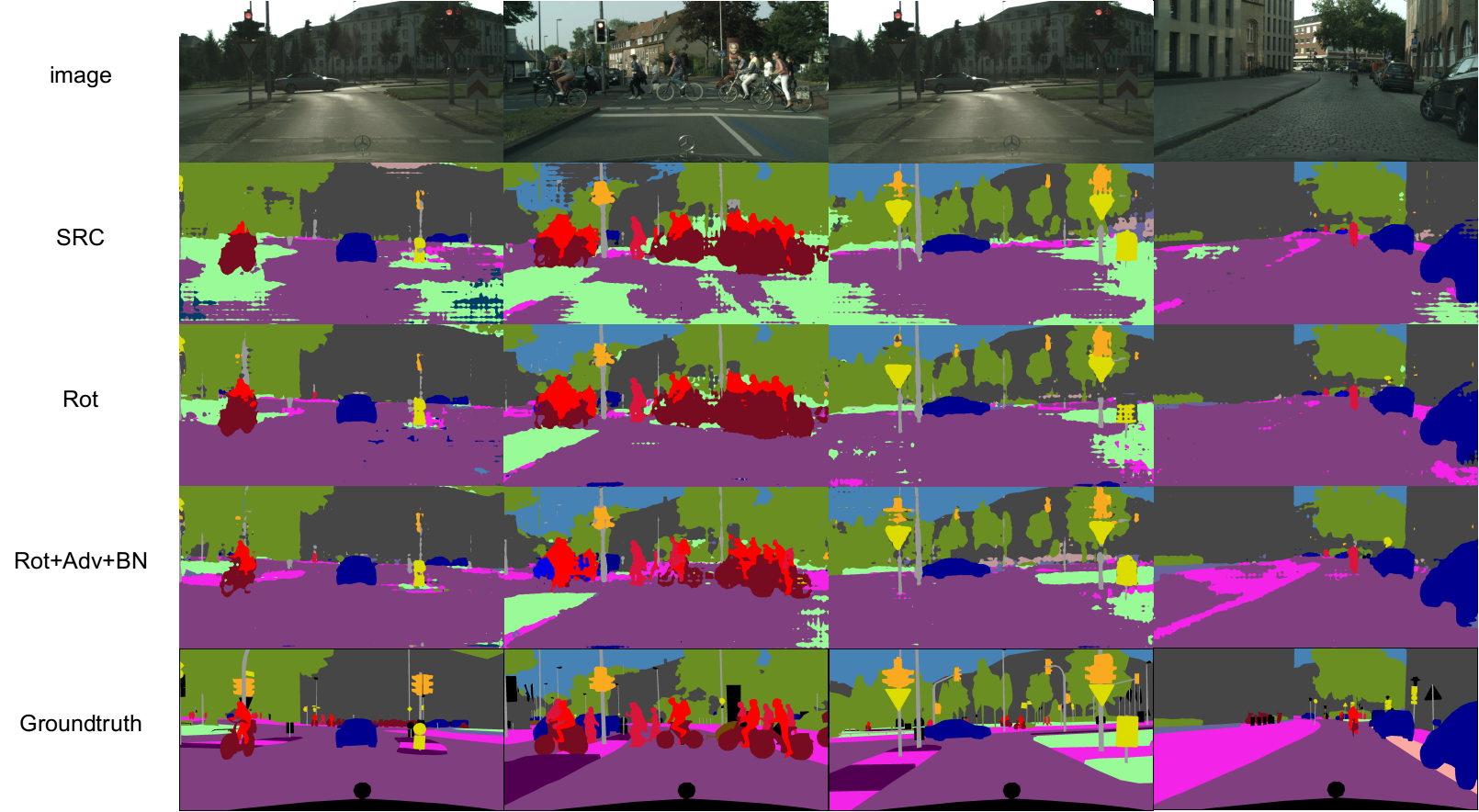}
\caption{Qualitative results of GTA2CS domain adaptation. The first row and last row are the input images and corresponding groundtruthes respectively. The second row shows results from source domain model. The third and fourth rows are results from the proposed Rot and Rot+Adv+BN models. ResNet-101 is used as backbone.}
\label{fig:qualitative}
\end{figure*}

\subsection{Comparison to the state-of-the-art}

Lastly, we compare our method to some recently published state-of-the-art works which use similar architectures to ours. The results are shown in Table~\ref{tab:sota}. The compared methods cover large varieties of domain adaptation mechanisms, including input/feature/output level alignment methods,  curriculum and self-labeling based methods. Some of these methods are also surveyed in \cite{curriculum:2018}. We refer the readers to \cite{curriculum:2018} for more details. The results in Table~\ref{tab:sota} show that our adapted models (\textbf{Adapt}) achieve comparable accuracies to the state-of-the-art. It is worth noting some of these state-of-the-art methods obtain worse results than we obtain when training with the source data alone
(SRC columns), so their relative gain is higher. 
On the other hand, with this work we aim at encouraging the use of pretext tasks for domain adaption of semantic segmentation models, which, as mentioned before, can be a complementary idea to others.
We also find that a deeper network (ResNet-101) can achieve better domain adaptation gain than the shallow one (DRN-26). 

\subsection{Discussion}
The experimental results reveal several insightful observations. (1) The current deep learning methods learn good feature representations for single domain but can not remove cross-domain discrepancy. (2) Using self-supervised representation learning can help to reduce domain shift, and the simplest image rotation prediction pretext task \textbf{Rot} can even achieve comparable performance to the state-of-the-art domain adaptation methods. (3) \textbf{Rot} turns out to be more robust than other alternatives, {\eg}, \textbf{MixRot}, \textbf{SPRot} and \textbf{Jigsaw}. (4) \textbf{Rot} is complementary to existing domain adaptation methods, {\eg}, adversarial based and batch normalization based ones. 

The reasons that self-supervised learning helps domain adaptation are as following: (1) the self-supervised learning involves source and target domain samples in a common supervised learning process which can help to learn cross-domain feature representations. This can be verified from the feature visualization on domain alignment, {\eg}, \figurename~\ref{fig:tsne-office} and \figurename~\ref{fig:tsne-pacs}. (2) Because the self-supervised learning and the main task are in a joint multi-task learning process, the model of the main task also learns from the cross-domain feature representations. As a result, the final model achieves domain adaptation on target domain.

Based on our experiments, we also have following findings about how to design a good self-supervised pretext task for domain adaptation: (1) As a common practice on many deep learning tasks, a deeper architecture can achieve better self-supervised domain adaptation performance than a shallow one. (2) Better performance on representation learning, better performance on domain adaptation, {\eg}, \textbf{Rot} vs \textbf{Jigsaw}. (3) More complex pretext task does not lead to better domain adaptation performance, {\eg}, \textbf{SPRot} is outperformed by \textbf{Rot}. (4) \textbf{Adv} and \textbf{Adv+BN} can further improve \textbf{Rot} if the performance of \textbf{Adv} is not worse than \textbf{Rot}. In this work, we only investigated several simple self-supervised learning strategies, we believe that for better self-supervised domain adaptation there are still large space to explore. 

\section{Conclusion}
\label{sec:conclusion}
In this work, we have explored self-supervised learning for domain adaptation. We have shown that a simple image rotation prediction (pretext task) self-supervision can achieve state-of-the-art domain adaptation performance. We have studied several pretext tasks as well as complementary domain adaptation strategies. Taking object recognition and semantic segmentation of urban scenes as relevant use cases, we have performed an ablative analysis of the different components included in our overall domain adaptation procedure. 
As future work, we would like to investigate more pretext tasks and to apply our method to other relevant vision tasks.


\section*{Acknowledgment}
This work is supported by the National Natural Science Foundation of China (NSFC, NO. 61601508, 61790565 and 61803380). Antonio M. L\'opez acknowledges the financial support by the Spanish project TIN2017-88709-R (MINECO/AEI/FEDER, UE), as well as the financial support by ICREA under the ICREA Academia Program. Antonio also thanks the Generalitat de Catalunya CERCA Program and its ACCIO agency. 

\bibliographystyle{IEEEtran}
\bibliography{references}

%

\vfill


\end{document}